\documentclass[10pt,letterpaper]{article}

\usepackage[T1]{fontenc}
\usepackage[utf8]{inputenc}
\usepackage{lmodern}
\usepackage[margin=1in]{geometry}
\usepackage{amsmath,amssymb}
\usepackage{graphicx}
\usepackage{float}
\usepackage{booktabs,longtable,array,calc}
\usepackage{caption}
\usepackage{enumitem}
\usepackage{xcolor}
\usepackage[numbers,sort&compress]{natbib}
\usepackage{url}
\usepackage{hyperref}

\definecolor{linkblue}{HTML}{1F5A94}
\hypersetup{
  colorlinks=true,
  linkcolor=linkblue,
  citecolor=linkblue,
  urlcolor=linkblue,
  pdftitle={TCAM for Autonomous Deformable Manipulation: The RMC² Champion System for WBCD 2026 Track 4},
  pdfauthor={Guangrui Shen, Zhili He, Shigang Wang, Yuanjun Sun, Qing Yu}
}

\setlist{leftmargin=*,itemsep=1pt,topsep=3pt}
\newcolumntype{P}[1]{>{\raggedright\arraybackslash}p{#1}}

\title{\vspace{-1.5em}\textbf{TCAM for Autonomous Deformable Manipulation: The RMC² Champion System for WBCD 2026 Track 4}}
\author{%
  \textbf{Guangrui Shen\textsuperscript{1}, Zhili He\textsuperscript{1}, Shigang Wang\textsuperscript{1}, Yuanjun Sun\textsuperscript{1}, Qing Yu\textsuperscript{1,*}}\\[0.35em]
  \normalsize \textsuperscript{1} TermiTech, Shenzhen, China\\[0.2em]
  \small \textsuperscript{*} Corresponding author. Email: \href{mailto:felix@termitech.cn}{felix@termitech.cn}
}
\date{\small Technical Report}

\begin{document}
\maketitle

\begin{abstract}
This technical report describes the RMC² Team's champion solution for the WBCD 2026 Track 4: Deformable Manipulation Challenge. The task requires a robot to pick a single T-shirt from a stack, load it onto a printing pallet, align the collar with a target area, and smooth the printing region, a sequence that involves single-layer separation, deformable transport, precise placement, and contact-rich surface adjustment. The competition strongly incentivizes fully autonomous execution, motivating the development of an autonomous solution. We built a fully autonomous system around the TCAM (TermiBrain Causal Action Model) framework, with the design principle that hardware, perception, data, and learning should jointly reduce the physical interaction complexity the policy must handle. A custom 3D-printed gripper designed for single-layer fabric separation improves picking reliability on a dual-arm ARX X5 platform. A wrist-centric four-camera setup pairs upper fisheye cameras for task-level context with lower RGB cameras for close-range gripper--cloth contact observation. We combine portable UMI-style demonstrations with real-robot demonstrations collected on the deployable platform to provide both broad manipulation priors and deployment-specific dynamics. TCAM ties these components into a closed loop: each trajectory is analyzed to identify the physical factors contributing to its outcome, driving targeted data recollection and policy fine-tuning. The policy outputs 30-step end-effector delta-pose action chunks from a multi-view VLA backbone. In the final competition, our system loaded 25 T-shirts at an average of approximately 23 seconds per attempt, with 22 achieving the required surface smoothness, securing first place in Track 4.
\end{abstract}

\vspace{0.75em}

\section{Introduction}\label{introduction}

Deformable-object manipulation remains one of the most difficult problems in robotics. Unlike rigid objects, whose states can be described by a compact pose, garments have high-dimensional, partially observable configurations that change continuously under gravity, friction, stretching, and robot contact. A small difference in grasp point can determine whether one or two layers are picked up; a slight change in release timing can turn a flat placement into a wrinkled one. These sensitivities make garment handling qualitatively harder than rigid pick-and-place.

The WBCD 2026 Track 4: Deformable Manipulation Challenge provides a concrete evaluation of this problem in a manufacturing-inspired setting \citep{wbcd2026rules}. The robot must pick a single T-shirt from a stack, load it onto a printing pallet, align the collar with a target area, and smooth the printing region. The scoring rules assign separate points for successful pick-and-load, collar alignment, and surface smoothness, with a multiplier that strongly favors fully autonomous execution over teleoperation. This makes the competition not only a test of manipulation capability but also a test of system-level autonomy.

Teleoperation is a natural baseline for such tasks. A human operator can interpret garment states, recover from unexpected deformations, and adapt strategies on the fly. However, relying on teleoperation forgoes the highest scoring multiplier and does not demonstrate deployable robotic intelligence. The alternative is to build a learned autonomous policy. Recent Vision-Language-Action (VLA) systems have demonstrated impressive generalization by connecting large vision-language backbones to robot action generation \citep{black2024pi0}, scaling to heterogeneous embodiments \citep{physicalintelligence2025pi05}, incorporating embodied memory for long-horizon reasoning \citep{torne2026mem}, and learning from real-world corrections \citep{physicalintelligence2025pi06}. However, applying VLA methods to deformable manipulation introduces specific challenges that general-purpose systems do not directly address. First, the contact state between gripper and fabric is local and easily occluded, making it difficult to observe from standard camera placements. Second, garment deformation is highly sensitive to small action variations, so the policy must operate on a narrow distribution of physically successful interactions. Third, a single round of data collection rarely covers the failure modes that emerge during deployment, requiring an iterative process of testing, diagnosing, and retraining.

Our approach addresses these challenges not by solving them independently, but by co-designing hardware, perception, data collection, and learning so that they jointly narrow the physical interaction distribution the policy must handle. A custom 3D-printed gripper makes single-layer picking more physically reliable, reducing the space of contact outcomes the policy encounters. A wrist-centric four-camera system provides both task-level context and close-range contact observation, making the critical gripper--cloth interaction state directly visible. A mixed data pipeline combining UMI-style and real-robot demonstrations gives the policy both broad manipulation priors and deployment-specific dynamics. The TCAM (TermiBrain Causal Action Model) framework ties these components into a closed-loop workflow: after each deployment round, trajectories are analyzed to identify the physical factors contributing to success or failure, and the findings drive targeted data recollection and policy fine-tuning.

Our main contributions are as follows:

\begin{itemize}
\item
  A fully autonomous system for T-shirt picking, loading, alignment, and smoothing that secured first place in the WBCD 2026 Track 4 Challenge.
\item
  A custom gripper with hard fingertips, a Velcro-assisted flipping surface, and inward-angled geometry for reliable single-layer fabric separation.
\item
  A contact-aware wrist-centric four-view perception design combining wide-angle fisheye and close-range RGB cameras.
\item
  A mixed UMI and real-robot data pipeline with prompt-conditioned view configuration and iterative quality filtering.
\item
  The TCAM framework linking causal trajectory analysis and trajectory memory to end-effector delta-pose action-chunk learning in a data closed loop.
\end{itemize}

\section{Challenge Setup}\label{challenge-setup}

\subsection{Task Description}\label{task-description}

The Track 4 challenge requires the robot to complete one T-shirt loading cycle through three steps: picking, loading, and alignment with surface-quality evaluation.

\begin{table}[!t]
\centering
\caption{Task stages and success criteria for the WBCD 2026 Track 4 challenge}
\small
\begin{tabular}{@{}P{0.17\textwidth}P{0.34\textwidth}P{0.43\textwidth}@{}}
\toprule
\textbf{Stage} & \textbf{Objective} & \textbf{Success Criteria} \\
\midrule
Step 1: Picking & Identify the topmost T-shirt from a stack and grasp exactly one T-shirt. & Exactly one T-shirt is securely grasped without disturbing the rest of the stack. \\
Step 2: Loading & Transport the grasped T-shirt to the printing pallet and place it onto the pallet surface. & The T-shirt is placed on the pallet and covers the effective printing area appropriately. \\
Step 3: Alignment and Surface Quality & Align the collar region with the target area and smooth the surface in the printing region. & The collar region is aligned with the target region, and the target printing area has acceptable wrinkle quality. \\
\bottomrule
\end{tabular}
\end{table}

The task is challenging because success is determined not only by whether the T-shirt is moved, but also by whether the correct number of garments is picked, whether the collar is aligned, and whether the surface is smooth enough for printing.

\subsection{Scoring Characteristics and Competition Objective}\label{scoring-characteristics-and-competition-objective}

The scoring rule strongly encourages full autonomy under the official Track 4 procedure \citep{wbcd2026rules}. Each completed T-shirt cycle receives base points for successful pick-and-load, alignment, and smooth surface quality, and the per-cycle score is multiplied by the autonomy level declared for that run. Fully autonomous execution receives the highest multiplier. The rules also penalize common failure cases, including picking multiple T-shirts, leaving dislocated unpicked T-shirts, or performing human manual intervention during the manipulation steps.

\begin{table}[!t]
\centering
\caption{Scoring items, penalties, and autonomy multipliers for the WBCD 2026 Track 4 challenge}
\small
\begin{tabular}{@{}P{0.23\textwidth}P{0.16\textwidth}P{0.55\textwidth}@{}}
\toprule
\textbf{Scoring Item} & \textbf{Points / Penalty} & \textbf{Meaning} \\
\midrule
Pick + Load & +5 & The robot successfully picks a single T-shirt and loads it onto the pallet. \\
Alignment & +2.5 & The T-shirt collar region is aligned in the target region. \\
Smooth Surface & +2.5 & The target printing region is smooth; this score applies only if alignment is achieved. \\
Autonomy Multiplier & x1 / x2 / x4 & On-site teleoperation, remote teleoperation, and fully autonomous execution receive different multipliers. \\
Multiple T-shirts Picked & -5 & More than one T-shirt is placed onto the pallet in a single attempt. \\
Dislocated T-shirts Left Unpicked & -0.5 each & Remaining T-shirts are dislocated and the team gives up recovering or picking them. \\
Human Manual Intervention During Steps 1-3 & -5 & Manual intervention during the manipulation steps incurs a penalty. \\
\bottomrule
\end{tabular}
\end{table}

This scoring design changes the technical objective. The optimal strategy is not only to maximize the success rate of a single T-shirt attempt, but also to maximize accumulated score within the limited competition time. Therefore, our system was designed around three goals:

\begin{itemize}
\item
  Maintain fully autonomous robot execution during each manipulation attempt.
\item
  Maximize the probability of successful single-shirt picking, alignment, and smoothness.
\item
  Avoid spending excessive time on low-probability recovery once an attempt has clearly entered an unfavorable state.
\end{itemize}

During the final competition, the manipulation policy itself was fully autonomous. Human operators did not teleoperate the robot during the pick-load-align-smooth process. However, at the run-management level, the team used a score-aware early-termination strategy: when an attempt showed a clear failure trend, the team stopped the robot execution and proceeded to reset rather than spending additional time on uncertain recovery. This strategy was used to maximize the total score rate under the competition rules.

\begin{figure}[!t]
\centering
\includegraphics[width=0.92\linewidth]{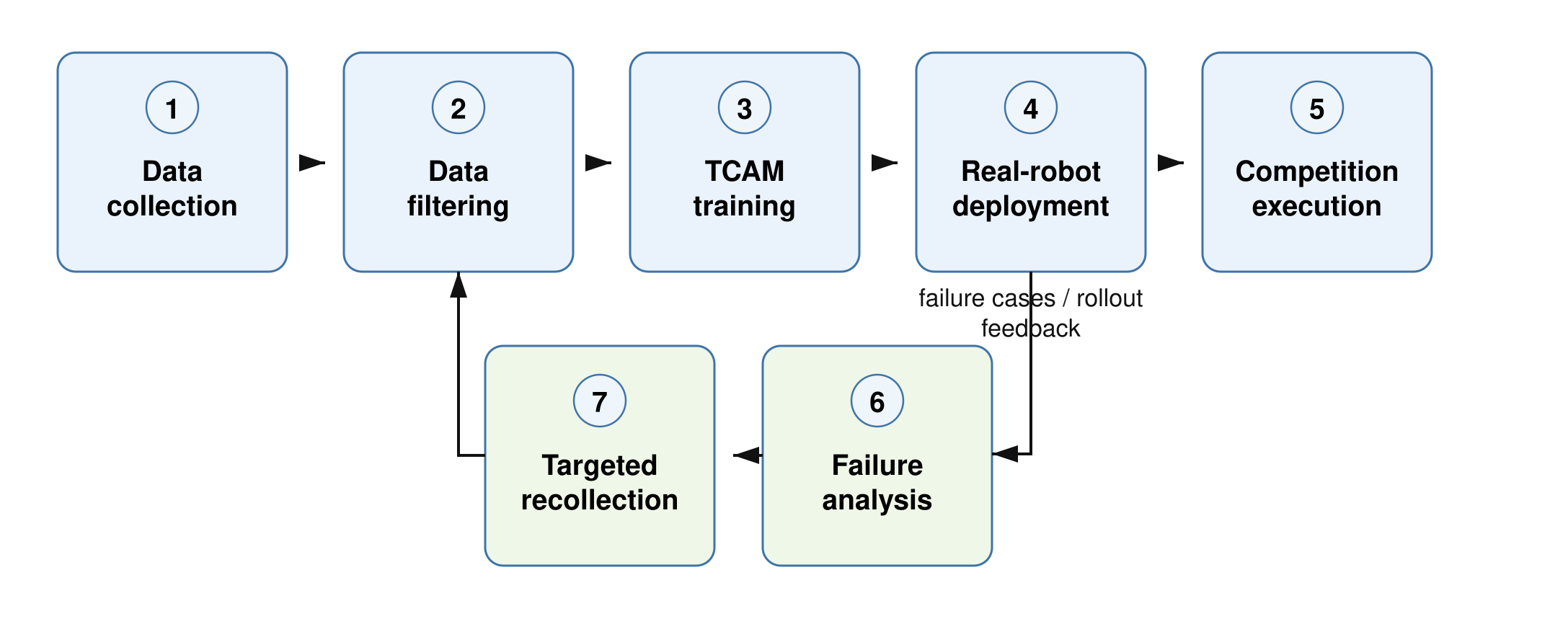}
\caption{Overall system pipeline with data closed-loop feedback. The system starts from mixed UMI-style and real-robot data collection, followed by filtering, TCAM training, real-robot deployment, competition execution, failure analysis, and targeted recollection. Failure cases and rollout feedback are stored in trajectory memory and fed back into iterative policy refinement.}
\label{fig:pipeline}
\end{figure}

\section{System Overview}\label{system-overview}

Figure~\ref{fig:pipeline} shows the overall system pipeline. The system integrates five components: a dual-arm ARX X5 robot platform (Section~\ref{robot-platform}), task-specific custom grippers (Section~\ref{custom-gripper-design}), a contact-aware wrist-centric four-view perception system (Section~\ref{multi-view-wrist-centric-perception}), a mixed UMI and real-robot data collection pipeline (Section~\ref{data-collection-and-data-closed-loop}), and the TCAM causal action learning framework (Section~\ref{tcam-termibrain-causal-action-model}).

These components are connected through a closed-loop data workflow. Rather than collecting a fixed dataset and training once, we repeatedly deploy the policy, analyze failure cases, collect targeted additional data, and retrain until the system reaches competition-level reliability. The TCAM trajectory memory stores both successful and failed rollouts together with their causal analysis results, enabling this iterative refinement cycle. Section~\ref{data-closed-loop} describes the closed-loop process, while Section~\ref{trajectory-memory-and-failure-driven-update} details the trajectory-memory mechanism.

\section{Hardware Design}\label{hardware-design}

\subsection{Robot Platform}\label{robot-platform}

We use a dual-arm ARX X5 robotic platform for the task. The two arms are positioned face-to-face to cover both the T-shirt stack and the printing pallet.

\subsection{Custom Gripper Design}\label{custom-gripper-design}

A key challenge is to reliably separate and grasp only the top T-shirt from a stack. Standard parallel grippers have difficulty flipping open the top fabric layer, and once lifted, fabric can slip through the closing fingertips, causing drop failures during transport.

To address this, we designed and 3D-printed task-specific grippers with three features:

\begin{itemize}
\item
  \textbf{Hard fingertip design.} The fingertips are rigid rather than soft-padded, improving the ability to pick and lift the top layer.
\item
  \textbf{Velcro-assisted top-layer flipping side.} One side of the gripper is equipped with hook-side Velcro to flip open the top layer of the stack, exposing a graspable region and increasing the likelihood of grasping only the uppermost T-shirt.
\item
  \textbf{Inward-angled gripper geometry.} The fingertips are angled slightly inward so that they meet more tightly upon closure, improving garment retention during lifting and transport.
\end{itemize}

This hardware design is not an isolated add-on; it changes the learning problem itself. By making single-layer picking physically easier and more repeatable, the gripper reduces the burden on the policy and improves the quality of collected demonstrations.

\begin{figure}[!t]
\centering
\includegraphics[width=0.47\linewidth]{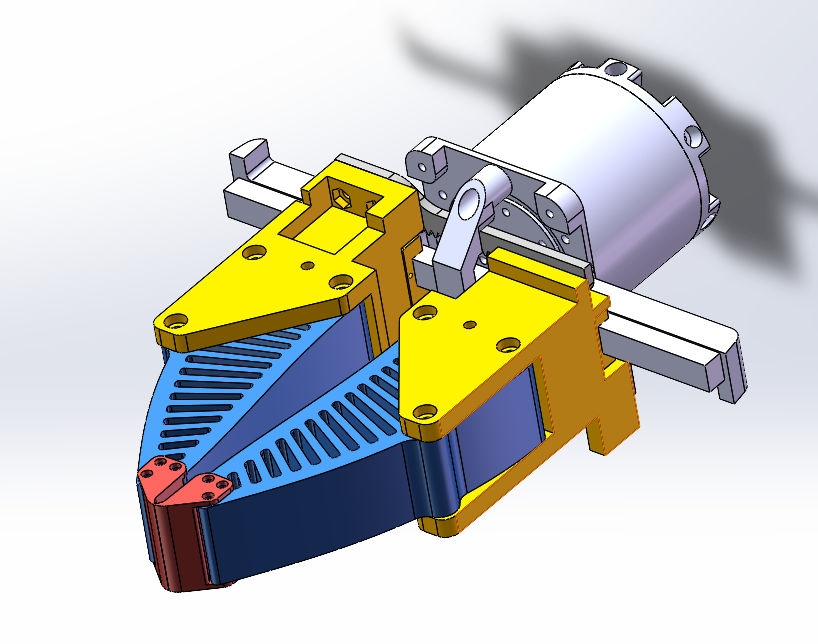}\hfill
\includegraphics[width=0.47\linewidth]{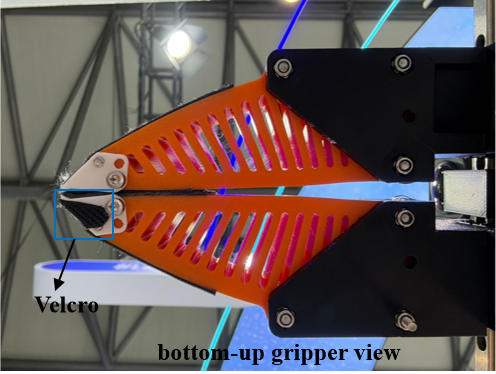}
\caption{Custom gripper design.}
\label{fig:gripper}
\end{figure}

\subsection{Printing Pallet Adaptation}\label{printing-pallet-adaptation}

We also adapted the printing pallet and workspace to reduce physical failure modes and improve visual observability.

The original pallet had rough wooden side edges that created friction during T-shirt loading, and protruding screws underneath could catch the fabric during insertion. We covered the side edges with adhesive tape to allow the fabric to slide more easily, and added a 3D-printed protective cover underneath the pallet to prevent the garment from being hooked by the screws.

To improve visual references for the policy, we added high-contrast black tape markers to the pallet surface and placed a patterned tablecloth under the workspace. The original white table surface was visually similar to the white T-shirts, making garment boundaries difficult to distinguish in camera observations. The textured tablecloth introduced stronger background contrast. Although we did not isolate the contribution of each modification through controlled ablations, the combined setup provided more stable visual cues for the TCAM policy in practice.

\begin{figure}[!t]
\centering
\includegraphics[width=0.82\linewidth]{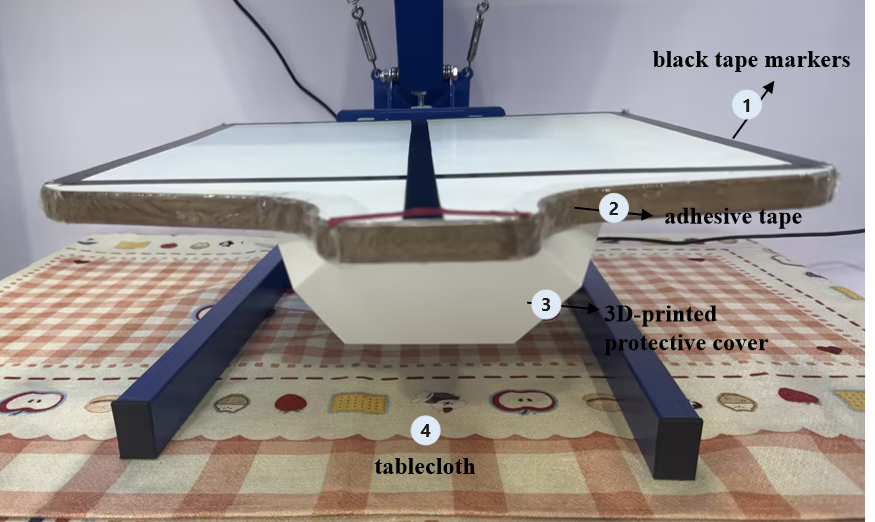}
\caption{Printing pallet and workspace adaptation. The rough wooden side edges of the pallet were covered with adhesive tape to reduce friction during T-shirt loading. A protective cover was added underneath the pallet to prevent the garment from being caught by protruding screws. High-contrast black tape markers on the pallet and a patterned tablecloth under the workspace were used to provide stronger visual references for the TCAM policy.}
\label{fig:pallet}
\end{figure}

\section{Multi-View Wrist-Centric Perception}\label{multi-view-wrist-centric-perception}

\subsection{Motivation}\label{motivation}

A single camera is insufficient for deformable T-shirt manipulation. The garment state is frequently occluded by the grippers, and some of the most critical information is local: whether the gripper has contacted only the top layer, whether the garment edge is being lifted. These states are difficult to infer from a fixed external camera or a single wrist-mounted view.

We therefore designed a four-camera wrist-mounted observation system. All cameras move with the robot arms, and the system is split into two complementary tiers: upper wrist fisheye cameras for wide-angle task context and lower wrist RGB cameras for close-range gripper--cloth contact observation.

\subsection{Camera Configuration}\label{camera-configuration}

Each robotic arm is equipped with two wrist cameras: an upper fisheye camera and a lower RGB camera, giving four camera streams in total. Table~3 summarizes the configuration.

\begin{table}[!t]
\centering
\caption{Wrist-mounted camera configuration and observation roles}
\small
\begin{tabular}{@{}P{0.16\textwidth}P{0.14\textwidth}P{0.14\textwidth}P{0.12\textwidth}P{0.335\textwidth}@{}}
\toprule
\textbf{View} & \textbf{Camera Type} & \textbf{Resolution} & \textbf{Frame Rate} & \textbf{Role} \\
\midrule
Left upper wrist & Fisheye RGB & 1920 x 1080 & 30 fps & Wide-angle context observation \\
Right upper wrist & Fisheye RGB & 1920 x 1080 & 30 fps & Wide-angle context observation \\
Left lower wrist & Normal RGB & 320 x 240 & 30 fps & Close-range gripper-cloth contact observation \\
Right lower wrist & Normal RGB & 320 x 240 & 30 fps & Close-range gripper-cloth contact observation \\
\bottomrule
\end{tabular}
\end{table}

All camera streams are resized to \(224 \times 224\) before being fed into the policy.

\subsection{Upper and Lower View Roles}\label{upper-and-lower-view-roles}

The upper fisheye cameras and lower RGB cameras serve distinct and complementary roles.

The upper fisheye views provide wide-angle context: the T-shirt stack, pallet position, arm configuration, and overall garment pose. These views indicate the current stage of the task and the garment's global configuration.

The lower RGB cameras provide direct visual evidence of the gripper--cloth interaction: the gripper opening relative to the cloth edge, whether the gripper is above or below the fabric, whether the top layer is being separated, and whether multiple layers are being unintentionally lifted. This local contact information is critical for single-layer picking. In deployment, we observed that the lower wrist views improved top-layer picking and reduced double-picking failures.

This separation of global context and local contact observation is the central design principle of our perception system.

\begin{figure}[!t]
\centering
\includegraphics[width=0.84\linewidth]{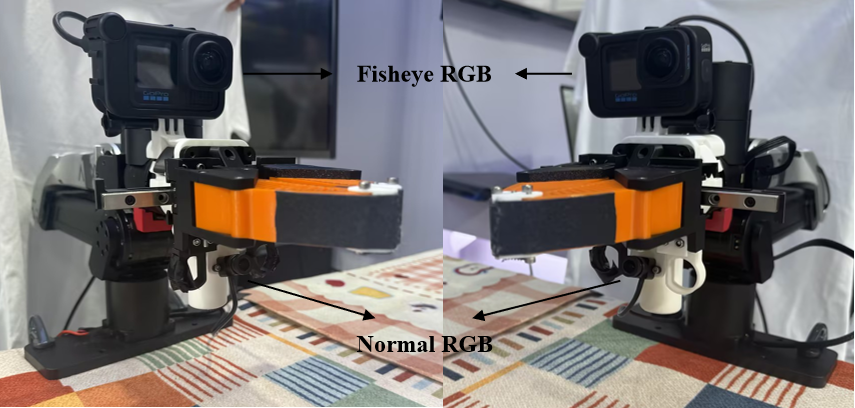}
\caption{Four-view wrist observation.}
\label{fig:wrist-cameras}
\end{figure}

\section{Data Collection and Data Closed Loop}\label{data-collection-and-data-closed-loop}

\subsection{Raw Demonstration Collection}\label{raw-demonstration-collection}

We collected demonstrations from two sources: UMI-style data collection (600 episodes) and real-robot data collection (500 episodes).

The UMI-style data follows the portable demonstration framework introduced in \citep{chi2024umi}. The UMI device is lightweight and easy to deploy across different scenes, table layouts, lighting conditions, and garment configurations, providing broader data coverage for scene generalization. However, our current UMI structure does not conveniently support mounting lower wrist cameras, so UMI episodes contain only the upper fisheye views.

Real-robot demonstrations were collected directly on the deployable ARX X5 platform with the full four-view camera setup, in both our practice environment and the competition field. These episodes capture the actual robot dynamics, custom gripper--fabric contact, control latency, and competition-specific workspace conditions. During training, prompts distinguish two-view UMI episodes from four-view real-robot episodes; at competition-time inference, the four-view prompt conditions the model toward the deployable rollout distribution.

Unlike conventional master-slave teleoperation, our real-robot data collection uses a master-deployment integrated setup: operators directly drag the deployable ARX X5 arms under force-compensation mode to record demonstrations. This substantially reduces the embodiment gap between data collection and final deployment.

\subsection{Final Training Dataset}\label{final-training-dataset}

After quality filtering, the final training dataset contains 600 mixed episodes: approximately 200 UMI-style and 400 real-robot. Table~4 summarizes the dataset composition and the primary contribution of each source.

\begingroup
\small
\begin{table}[!t]
\centering
\caption{Mixed UMI-style and real-robot demonstration dataset with prompt-conditioned view configurations}
\begin{tabular}{@{}P{0.16\textwidth}P{0.10\textwidth}P{0.17\textwidth}P{0.45\textwidth}@{}}
\toprule
\textbf{Data Source} & \textbf{\shortstack[l]{Raw\\Episodes}} & \textbf{\shortstack[l]{Final Training\\Episodes}} & \textbf{Primary Contribution}\\
\midrule
UMI-style collection & 600 & $\sim$200 & Portable cross-scene demonstrations for human manipulation priors, broader scene coverage, and robustness; upper wrist fisheye views only due to UMI mounting constraints.\\
Real-robot collection & 500 & $\sim$400 & In-practice and competition-field four-view demonstrations collected directly on the deployable ARX X5 under force-compensation dragging, capturing contact views, robot dynamics, gripper behavior, latency, and camera geometry.\\
Final mixed dataset & 1100 raw total & 600 & Prompt-conditioned mixture of two-view UMI data and four-view deployable real-robot data for competition-time inference.\\
\bottomrule
\end{tabular}
\end{table}
\endgroup

The filtering process removes low-quality trajectories, ambiguous demonstrations, poor camera synchronization, and episodes with undesirable outcomes. This step is especially important for deformable-object learning, where inconsistent trajectories can teach the policy unstable contact behavior.

\subsection{Data Closed Loop}\label{data-closed-loop}

Our training follows an iterative closed-loop workflow rather than a single-round data collection and training process:

\begin{itemize}
\item
  Collect initial UMI-style and real-robot demonstrations.
\item
  Train the TCAM policy.
\item
  Deploy the policy on the real robot and record rollout outcomes.
\item
  Analyze failure modes to identify their physical causes.
\item
  Collect targeted additional data for the identified failure cases.
\item
  Filter and rebalance the dataset, then retrain and redeploy.
\end{itemize}

The most common failure modes observed during this process include missed top-edge grasping, double-picking, garment loss during transport, poor initial pallet placement, collar misalignment, remaining wrinkles in the printing region, and over-adjustment that worsens an otherwise acceptable placement.

This iterative process allows the system to improve in the exact failure distribution of the competition task. The trajectory-level analysis and memory mechanism that supports this loop is described in Section~\ref{trajectory-memory-and-failure-driven-update}.

\section{TCAM: TermiBrain Causal Action Model}\label{tcam-termibrain-causal-action-model}

\subsection{Framework Overview}\label{tcam-framework-overview}

TCAM (TermiBrain Causal Action Model) is our system-level learning framework built around a self-developed multi-view Vision-Language-Action policy.

Rather than treating demonstrations solely as trajectories for imitation, TCAM organizes each rollout as a causal interaction chain linking visual observation and gripper--cloth contact to action execution, garment deformation, and task outcome. TCAM analyzes this chain through a semi-automated workflow rather than fully automated causal attribution. Software first aggregates synchronized observations, actions, and outcome labels into a structured rollout record; human reviewers then inspect the record, identify physical factors such as missed edge contact, double picking, transport slip, collar misalignment, remaining wrinkles, or over-adjustment, and confirm the final causal tags. This analysis connects what the robot perceived and executed to how the garment responded and why the task succeeded or failed.

\subsection{Trajectory Memory}\label{trajectory-memory-and-failure-driven-update}

Successful trajectories, failed trajectories, and the human-verified causal analysis results produced by this semi-automated workflow are stored in a trajectory memory buffer. This memory supports three functions: replay for retraining, identification of failure modes for targeted data recollection, and dataset rebalancing toward difficult cases. The iterative deploy--analyze--recollect--retrain loop described in Section~\ref{data-closed-loop} is driven by the contents of this memory.

\subsection{Policy Input and Output}\label{policy-input-and-action-output}

The policy receives synchronized wrist-camera images resized to \(224 \times 224\), robot proprioception---including end-effector poses and gripper states---and task-conditioning prompts. As described in Section~\ref{raw-demonstration-collection}, prompts distinguish two-view UMI data from four-view real-robot data; competition-time inference uses the four-view prompt.

The policy outputs 30-step chunks of end-effector delta-pose actions for both arms. Action chunking \citep{zhao2023act} improves temporal consistency and reduces frame-to-frame corrections that can create wrinkles, shift the collar, or cause the cloth to slip.

\section{Autonomous Execution and Run Management}\label{autonomous-execution-and-run-management}

During every active manipulation attempt, all robot motion is generated autonomously. No human teleoperation or corrective action is used during picking, loading, alignment, or smoothing. The policy receives wrist-camera observations and outputs delta-pose action chunks as described in Section~\ref{policy-input-and-action-output}, enabling the robot to execute the complete pick--load--align--smooth cycle without human motion commands.

At the run-management level, the team adopted the score-aware early-termination strategy described in Section~\ref{scoring-characteristics-and-competition-objective}. Concretely, if an attempt showed clear signs that recovery was unlikely within the remaining time, such as a failed grasp, severe misplacement, or a difficult wrinkle configuration, the team stopped the robot and reset the workspace rather than allowing the policy to continue on a low-probability recovery. The termination decision was based on human monitoring during the competition rather than an autonomous failure detector.

This reflects a practical distinction worth noting: the manipulation policy itself was fully autonomous, while the decision to continue or terminate an attempt remained a human supervisory decision. Integrating this decision into the autonomous system is discussed as future work in Section~\ref{discussion}.

\section{Competition Performance}\label{competition-performance}

In the final competition, our system completed 25 T-shirts. Among them, 22 were judged as smooth in the target printing region. Table~5 summarizes the results.

\begin{table}[!t]
\centering
\caption{Final performance summary in the WBCD 2026 Track 4 challenge}
\small
\begin{tabular}{@{}P{0.40\textwidth}P{0.54\textwidth}@{}}
\toprule
\textbf{Metric} & \textbf{Result} \\
\midrule
Completed T-shirts & 25 \\
Smooth T-shirts & 22 \\
Average time per attempt & Approximately 23 seconds \\
Unused autonomous loading time & Approximately 15 seconds \\
Manipulation control mode & Fully autonomous TCAM policy execution \\
\bottomrule
\end{tabular}
\end{table}

Counting both successful and early-terminated attempts, the average time per T-shirt attempt was approximately 23 seconds. The session followed the official 30-minute schedule with an additional one-minute time compensation due to judging-time conditions. Even under this setting, the system still had approximately 15 seconds of autonomous loading time unused.

Post-competition analysis showed that alignment-specific demonstrations were underrepresented in the training dataset. During the competition, only a limited number of attempts received full alignment scores. Future work will expand alignment-focused data collection and evaluate its effect through targeted retraining.

\section{Discussion}\label{discussion}

Our competition experience suggests several practical observations for deformable manipulation system design. Task-specific hardware and learned policy should be co-designed: the custom gripper did not merely assist the policy but changed the physical interaction distribution it needed to learn (Section~\ref{custom-gripper-design}). For perception, the value of the four-camera setup lies not in the number of views but in contact observability, particularly the lower wrist cameras that make gripper--cloth interaction directly visible (Section~\ref{upper-and-lower-view-roles}). UMI-style and real-robot data serve complementary roles: UMI provides portable cross-scene coverage while real-robot episodes capture deployment-specific dynamics and contact views (Section~\ref{raw-demonstration-collection}). Finally, a single round of data collection is rarely sufficient; iterative deployment, failure analysis, and targeted recollection were necessary to reach competition-level reliability (Section~\ref{data-closed-loop}).

Several limitations remain. Although the policy performs well in the competition setting, long-horizon recovery from severe failures remains challenging. The early-termination decision was made by human observation rather than an autonomous failure detector (Section~\ref{autonomous-execution-and-run-management}). A fully integrated system should automatically estimate whether the current state is recoverable and decide whether to continue or abort based on expected score and remaining time. Future work will also focus on scaling the data closed loop, improving generalization across textile categories, and extending the system toward manufacturing-oriented garment handling.

\section{Conclusion}\label{conclusion}

We presented the RMC² Team's champion solution for the WBCD 2026 Track 4: Deformable Manipulation Challenge. Our TCAM-based system autonomously executes all robot motion for T-shirt picking, loading, alignment, and smoothing, combining task-specific gripper design, wrist-centric four-view perception, mixed UMI and real-robot data, and closed-loop policy refinement. In the final competition, the system completed 25 T-shirts, 22 of which met the required surface-smoothness criterion, and secured first place in Track 4.

\section*{Acknowledgements}\label{acknowledgements}
\addcontentsline{toc}{section}{Acknowledgements}

We thank the ICRA WBCD 2026 organizers, judges, hardware providers, and volunteers for supporting the Track 4: Deformable Manipulation Challenge. We also thank all members of the RMC² Team and TermiTech for their work on hardware design, data collection, model training, system integration, and competition deployment.

\bibliographystyle{ieeetr}
\bibliography{references}

\end{document}